\documentclass[runningheads]{llncs}
\usepackage[T1]{fontenc}
\usepackage{graphicx}
\usepackage{amsmath}
\usepackage{amssymb}
\usepackage{multirow}
\usepackage{subfigure}
\usepackage{url}
\begin{document}
\title{A Scalable Benchmark Test Suite for Dynamic Multi-Objective Optimization with a Changing Number of Objectives}
\titlerunning{A Scalable Benchmark Test Suite for DMOP with a Changing NObjs}
%
\author{Ke Shang\inst{1,2} \and
Zhiyun Xiao\inst{1,2} \and
Yuxuan Liu\inst{3} \and
Jianguo Li\inst{4} \and
Shaojiang Wang\inst{4} \and
Wei Sun\inst{4}}
\authorrunning{K. Shang et al.}
%
\institute{School of Artificial Intelligence, Shenzhen University, Shenzhen 518060, China \and
National Engineering Laboratory for Big Data System Computing Technology, Shenzhen University, Shenzhen 518060, China\\
\email{kshang@foxmail.com} \and
Tsinghua University, Beijing 100084, China \and
Shenzhen ZTE Software Co., Ltd.}
\maketitle              
\begin{abstract}
Dynamic multi-objective optimization with a changing number of objectives has recently attracted increasing attention due to its relevance to real-world problems whose evaluation criteria may evolve over time.
However, existing benchmark test suites for this problem setting suffer from a fundamental limitation: when the number of objectives changes, the objective functions themselves also change implicitly.
This makes it difficult to isolate and evaluate an algorithm's capability to handle dynamics in the number of objectives alone.
In this paper, we analyze this issue in detail and show that several theoretical properties claimed in prior studies rely on an assumption that is violated by commonly used test suites.
To address this problem, we propose a scalable benchmark test suite in which the objective functions are fixed throughout the optimization process, while the number of active objectives changes over time.
Our benchmark is constructed by defining a maximum-objective problem and dynamically selecting subsets of objectives.
To avoid degeneracy issues in classical DTLZ and WFG problems, we adopt Minus-DTLZ and Minus-WFG formulations, in which all objectives are mutually conflicting.
Extensive benchmark studies using representative algorithms from the literature demonstrate the usefulness and flexibility of the proposed test suite.
\end{abstract}
\section{Introduction}

Multi-objective optimization problems (MOPs) involve the simultaneous optimization of multiple conflicting objectives.
In many real-world applications, optimization takes place in dynamic environments, where problem characteristics may change over time.
Dynamic multi-objective optimization (DMO) aims to design algorithms that can effectively track and adapt to such changes~\cite{farina2004dynamic}.

Among various forms of dynamics, one particularly challenging and practically relevant scenario is \emph{dynamic multi-objective optimization with a changing number of objectives} (DMO-CNO).
In this setting, the number of objectives varies over time, for example due to changing decision requirements, evolving stakeholder preferences, or the gradual refinement or abstraction of evaluation criteria.
Such a setting combines the difficulties of dynamic optimization and many-objective optimization, both of which have been recognized as challenging research directions in evolutionary computation~\cite{jiang2023survey,chand2015quick,ishibuchi2008many}.
Related optimization problems also arise in communication and networking systems, where large search spaces, multiple criteria, and changing operating conditions often need to be handled jointly~\cite{liao2017evolutionary,zhang2021handover}.

Formally, a dynamic multi-objective optimization problem with a changing number of objectives can be defined as
\begin{equation}
\min_{\mathbf{x} \in \Omega} \ \mathbf{F}(\mathbf{x}, t) = \left( f_1(\mathbf{x}), f_2(\mathbf{x}), \dots, f_{m(t)}(\mathbf{x}) \right),
\end{equation}
where $\mathbf{x}$ is the decision vector, $t$ denotes time, and $m(t)$ is the number of objectives at time $t$.
In this paper, we focus exclusively on dynamics in the \emph{number of objectives}.
That is, we assume that the objective functions $f_i(\cdot)$ themselves do not change over time.

Several studies have investigated DMO-CNO in recent years.
Chen \emph{et al.}~\cite{chen2017dynamic} first formalized this problem setting and proposed evolutionary algorithms to cope with expanding and contracting Pareto fronts.
Subsequent works explored knowledge transfer~\cite{ruan2024knowledge}, Pareto set expansion and contraction~\cite{ruan2024learning}, and severely changing numbers of objectives~\cite{ruan2025coping}.

Despite these algorithmic advances, the benchmark test suites used in existing studies remain problematic.
Most existing benchmarks are constructed by directly modifying classical static test problems, such as DTLZ or WFG~\cite{deb2005dtlz,huband2005wfg}, by replacing the number of objectives $m$ with a time-dependent function $m(t)$ in the problem formulation.
However, since the objective functions themselves depend on $m$, this approach implicitly changes the definitions of the objective functions over time.

This leads to a conceptual inconsistency: although the problem is described as one where only the number of objectives changes, in reality both the number and the definitions of objectives vary.
What we desire instead is a benchmark where objectives are added or removed over time \emph{without altering the remaining objective functions}.
This distinction is important because benchmark design strongly affects what algorithmic capabilities are actually being evaluated in dynamic optimization studies~\cite{nguyen2012survey,jiang2023survey}.

To address this issue, we propose a scalable benchmark test suite for DMO-CNO.
Our benchmark is constructed by first defining a problem with a maximum number of objectives, and then dynamically selecting subsets of objectives to form the active problem at each time step.
In this way, objectives can be added or removed without modifying any existing objective functions.
Furthermore, by adopting Minus-DTLZ and Minus-WFG formulations, we ensure that all objectives are mutually conflicting, avoiding degeneracy issues inherent in classical formulations.

The main contributions of this paper are summarized as follows:
\begin{itemize}
\item We identify and analyze a fundamental limitation of existing benchmark test suites for DMO-CNO.
\item We propose a scalable and flexible benchmark framework in which only the number of objectives changes over time.
\item We construct concrete test instances based on Minus-DTLZ and Minus-WFG problems.
\item We conduct benchmark studies using representative algorithms from the literature to demonstrate the effectiveness of the proposed test suite.
\end{itemize}

\section{Limitations of Existing Test Suites}

In~\cite{chen2017dynamic}, dynamic multi-objective optimization with a changing number of objectives is formulated by directly replacing the static number of objectives $m$ with a time-dependent function $m(t)$ in classical benchmark problems.
A similar approach is also adopted in~\cite{jiang2019scalable}.

Under this formulation, the objective vector at time $t$ is written as
\begin{equation}
\mathbf{F}(\mathbf{x}, t) = \left( f_1(\mathbf{x}, t), \dots, f_{m(t)}(\mathbf{x}, t) \right),
\end{equation}
where each objective explicitly depends on $t$.

For example, the F1 problem defined in~\cite{chen2017dynamic}, which is based on DTLZ1, is written as follows:
\begin{equation}
\begin{aligned}
\text{F1: } \quad
&f_1(\mathbf{x}, t) = (1 + g(\mathbf{x}, t))^{0.5}
\prod_{i=1}^{m(t)-1} x_i,\\
&f_{j = 2, \ldots, m(t)-1}(\mathbf{x}, t) = (1 + g(\mathbf{x}, t))^{0.5}
\left( \prod_{i=1}^{m(t)-j} x_i \right)
\left( 1 - x_{m(t)-j+1} \right), \\
&f_{m(t)}(\mathbf{x}, t) = (1 + g(\mathbf{x}, t))^{0.5}
\left( 1 - x_1 \right),
\end{aligned}
\end{equation}
where
\begin{equation}
g(\mathbf{x}, t)
=
100\left[
n - m(t) + 1
+
\sum_{i=m(t)}^{n}
\left(
(x_i - 0.5)^2
-
\cos\left(20\pi(x_i - 0.5)\right)
\right)
\right],
\end{equation}
and
\begin{equation}
x_i \in [0,1], \quad i = 1, \ldots, n.
\end{equation}

However, this formulation has an important drawback.
When the number of objectives changes from $m(t)$ to $m(t+1)$, not only are objectives added or removed, but the definitions of all existing objective functions are also modified (e.g., $f_1(\mathbf{x}, t)\neq f_1(\mathbf{x}, t+1)$ if $m(t)\neq m(t+1)$).
As a result, the optimization problem at time $t+1$ is not a simple expansion or contraction of the problem at time $t$, but an entirely new problem.

Meanwhile, the following theorem is established in~\cite{chen2017dynamic}:

\textbf{Theorem 1}~\cite{chen2017dynamic}:  
\textit{When increasing the number of objectives, the Pareto front (PF) and Pareto set (PS) at time step $t$ are subsets of those at time step $t+1$.
Conversely, when decreasing the number of objectives, the PF and PS at time step $t$ are supersets of those at time step $t+1$.}

This theorem implicitly assumes that objective functions remain unchanged over time and that only the number of objectives varies.
However, under the commonly used benchmark construction where objective functions depend on $m(t)$, this assumption is violated.
Consequently, the theorem may not hold for the benchmark problems defined in~\cite{chen2017dynamic} and~\cite{jiang2019scalable}.

This inconsistency can be seen directly from the DTLZ1-based F1 example above.
Consider $n=3$ and $\mathbf{x}=(0.5,0.5,0.5)$.
When $m(t)=2$, we have
\[
g(\mathbf{x}, t)=100\!\left[2+\sum_{i=2}^{3}\left((x_i-0.5)^2-\cos(20\pi(x_i-0.5))\right)\right]=0,
\]
and thus
\[
f_1(\mathbf{x}, t)=(1+g)^{0.5}x_1=0.5.
\]
When the number of objectives changes to $m(t+1)=3$, the same decision vector yields
\[
g(\mathbf{x}, t+1)=100\!\left[1+\sum_{i=3}^{3}\left((x_i-0.5)^2-\cos(20\pi(x_i-0.5))\right)\right]=0,
\]
but now
\[
f_1(\mathbf{x}, t+1)=(1+g)^{0.5}x_1x_2=0.25.
\]
Therefore, even an objective that remains active before and after the change is redefined by the benchmark itself.
The PF/PS inclusion relation in Theorem~1 cannot be inferred for such a construction, because the comparison is no longer between two problems sharing the same objective functions.

This analysis highlights the need for a benchmark test suite in which the objective functions are fixed throughout the optimization process and only the number of active objectives changes over time.

\section{Proposed Test Suite}

\subsection{General Framework}

The proposed test suite is constructed as follows.
First, we define a multi-objective optimization problem with a maximum number of objectives $m_{\max}$:
\begin{equation}
\mathbf{F}_{\max}(\mathbf{x}) = \left( f_1(\mathbf{x}), f_2(\mathbf{x}), \dots, f_{m_{\max}}(\mathbf{x}) \right).
\end{equation}

At each time step $t$, a subset of objectives
\begin{equation}
\mathcal{I}(t) \subseteq \{1,2,\dots,m_{\max}\}
\end{equation}
is selected to form the active optimization problem:
\begin{equation}
\mathbf{F}(\mathbf{x}, t) = \left( f_i(\mathbf{x}) \mid i \in \mathcal{I}(t) \right).
\end{equation}

When transitioning from time $t$ to $t+1$, objectives can be added to or removed from $\mathcal{I}(t)$.
Importantly, the definitions of all objective functions $f_i(\cdot)$ remain unchanged throughout the optimization process.

The time variable $t$ is defined in the same manner as in~\cite{chen2017dynamic}, i.e., $t=\lfloor\tau/\tau_t\rfloor$, which is controlled by the iteration counter $\tau$ and a predefined frequency of change $\tau_t$.

\subsection{Choice of Base Problems}

Following~\cite{deb2005dtlz,huband2005wfg}, we also adopt the widely used DTLZ and WFG test suites as base problems.
However, as pointed out in~\cite{ishibuchi2017performance}, classical DTLZ and WFG problems suffer from a degeneracy issue:
if $m_{\max}$ objectives are defined, any subset of $m_{\max}-1$ objectives is non-conflicting and can be optimized simultaneously, resulting in a single-point Pareto front.

\begin{figure*}[t]
\centering
    \includegraphics[width=\textwidth]{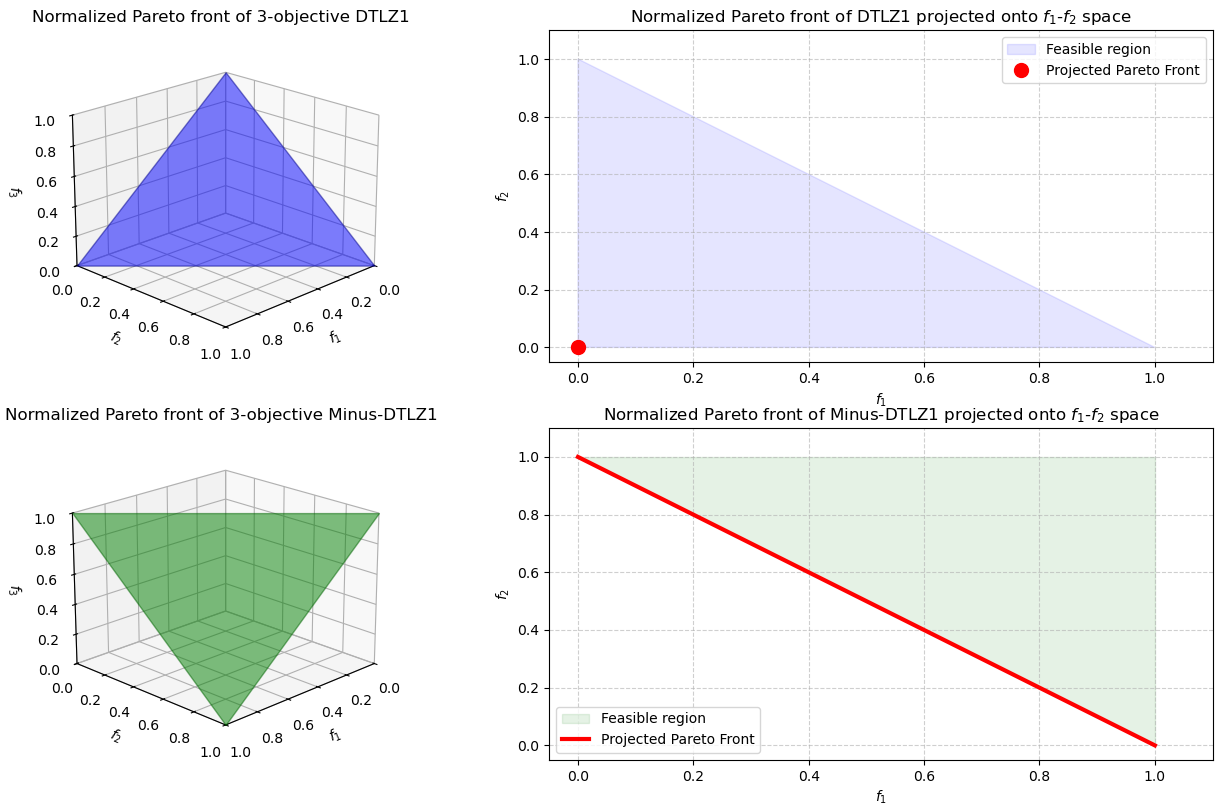}
\caption{Illustration of normalized Pareto fronts of 3-objective DTLZ1 and Minus-DTLZ1 and their projected Pareto fronts in $f_1$-$f_2$ space.}
\label{fig:pf}
\end{figure*}

For example, for an $m_{\max}=3$ DTLZ1 problem, if we choose $f_1$ and $f_2$ to form a two-objective problem, its Pareto front is a single point $(0,0)$ since $f_1$ and $f_2$ can be simultaneously minimized by letting the first decision variable $x_1=0$, as illustrated in Fig. \ref{fig:pf}. 

To avoid this issue, we adopt the Minus-DTLZ and Minus-WFG formulations proposed in~\cite{ishibuchi2017performance} (i.e., a minus sign is added to each objective function), in which all objectives are mutually conflicting.
This ensures that meaningful Pareto fronts exist for any subset of objectives, as illustrated in Fig. \ref{fig:pf}.

\subsection{Illustration of Dynamics}

The dynamics of objectives are specified by a sequence of index subsets $\mathcal{I}(t),t=1,2,...$.
For example, the sequence
\[
\{1,2,5\}, \ \{1,2,3,5\}, \ \{2,3,5\}
\]
defines three stages.
The first stage involves objectives $\{f_1,f_2,f_5\}$.
In the second stage, objective $f_3$ is added.
In the third stage, objective $f_1$ is removed.

This mechanism allows highly flexible and interpretable control over the dynamics of the number of objectives, which is desirable for benchmark construction in dynamic multi-objective optimization~\cite{farina2004dynamic,jiang2019scalable}.

\section{Experimental Study}

This section presents the experimental study conducted to evaluate the effectiveness of the proposed benchmark test suite for dynamic multi-objective optimization with a changing number of objectives.
Four representative algorithms from the literature are tested under different dynamic settings.
All experiments are conducted under identical conditions to ensure fair comparisons.

\subsection{Problem Settings}

Based on the proposed benchmark construction, we consider three dynamic settings with different degrees of objective-number variation.
All problems are constructed using the proposed Minus-DTLZ / Minus-WFG framework with a maximum number of objectives $m_{\max}$.
At each environmental change, the active objective set is updated according to a predefined sequence.

\subsubsection{Setting I: Mild Change (One Objective Added or Removed)}

The first setting simulates mild dynamics, where only one objective is added or removed at each change. We set $m_{\max}=6$ in this setting.
The number of objectives evolves as
\[
2 \rightarrow 3 \rightarrow 4 \rightarrow 5 \rightarrow 6 \rightarrow 5 \rightarrow 4 \rightarrow 3 \rightarrow 2.
\]

The corresponding objective index sequence is
\[
\{2,4\}, \{2,4,5\}, \{1,2,4,5\}, \{1,2,4,5,6\}, \{1,2,3,4,5,6\},
\]
\[
\{2,3,4,5,6\}, \{2,3,4,5\}, \{2,3,5\}, \{3,5\}.
\]

This setting evaluates the algorithms' ability to adapt to gradual expansions and contractions of the objective space.

\subsubsection{Setting II: Moderate Change (Two Objectives Added or Removed)}

The second setting simulates moderate dynamics, where two objectives are added or removed at each change. We set $m_{\max}=10$ in this setting.
The number of objectives evolves as
\[
2 \rightarrow 4 \rightarrow 6 \rightarrow 8 \rightarrow 10 \rightarrow 8 \rightarrow 6 \rightarrow 4 \rightarrow 2.
\]

The corresponding objective index sequence is
\[
\{2,7\}, \{2,5,7,10\}, \{1,2,5,6,7,10\},\{1,2,4,5,6,7,9,10\},\{1,2,3,4,5,6,7,8,9,10\},
\]
\[
\{1,2,3,5,6,8,9,10\}, \{2,3,5,6,9,10\}, \{2,5,6,9\}, \{5,6\}.
\]

This setting evaluates the robustness of algorithms under more pronounced objective changes.

\subsubsection{Setting III: Severe Change (Irregular Objective Variations)}

The third setting simulates severely changing numbers of objectives, where both the magnitude and direction of change vary irregularly. We set $m_{\max}=10$ in this setting.
The number of objectives evolves as
\[
2 \rightarrow 5 \rightarrow 10 \rightarrow 6 \rightarrow 3 \rightarrow 8 \rightarrow 4 \rightarrow 7 \rightarrow 9.
\]

The corresponding objective index sequence is
\[
\{3,8\}, \{2,3,6,7,8\}, \{1,2,3,4,5,6,7,8,9,10\},\{1,3,5,6,7,10\}, \{3,7,8\},
\]
\[
\{1,3,4,5,6,7,8,9\}, \{2,5,7,10\},\{1,2,4,5,6,9,10\}, \{1,2,3,4,5,6,7,8,10\}.
\]

This setting is designed to test the algorithms' adaptability under highly unstable objective dynamics.

Table~\ref{tab:dynamic_settings} summarizes the three dynamic settings considered in our experiments.

\begin{table}[t]
\caption{Summary of Dynamic Objective Settings}
\label{tab:dynamic_settings}
\centering
\begin{tabular}{c c c}
\hline
Setting & Objective Change Pattern & Change Severity \\
\hline
I & $2 \rightarrow 3 \rightarrow 4 \rightarrow 5 \rightarrow 6 \rightarrow 5 \rightarrow 4 \rightarrow 3 \rightarrow 2$ & Mild \\
II & $2 \rightarrow 4 \rightarrow 6 \rightarrow 8 \rightarrow 10 \rightarrow 8 \rightarrow 6 \rightarrow 4 \rightarrow 2$ & Moderate \\
III & Irregular ($2 \rightarrow 5 \rightarrow 10 \rightarrow 6 \rightarrow 3 \rightarrow 8 \rightarrow 4 \rightarrow 7 \rightarrow 9$) & Severe \\
\hline
\end{tabular}
\end{table}

\subsection{Compared Algorithms}

Four state-of-the-art algorithms for dynamic multi-objective optimization with a changing number of objectives are selected for comparison.

\textbf{DTAEA}~\cite{chen2017dynamic} is a dynamic two-archive evolutionary algorithm that maintains a convergence archive and a diversity archive.
It is specifically designed to handle the expansion and contraction of the Pareto front when the number of objectives changes.

\textbf{KTDMOEA}~\cite{ruan2024knowledge} incorporates a knowledge transfer mechanism to reuse useful information from previous environments.
By transferring knowledge across different objective dimensions, it aims to accelerate adaptation after objective changes.

\textbf{LEC}~\cite{ruan2024learning} employs a learning-based strategy to explicitly model the expansion and contraction of Pareto sets.
It leverages historical population information to guide the search process when objectives are added or removed.

\textbf{STA}~\cite{ruan2025coping} is designed to cope with severely changing numbers of objectives.
It introduces a specialized adaptation mechanism to handle large and irregular objective changes effectively.

All algorithms are implemented using the parameter settings recommended in their original publications.

\subsection{Parameter Settings}

The common experimental settings are summarized as follows:
\begin{itemize}
\item Population size is set to 300 for all algorithms.
\item Three change frequencies $\tau_t$ are considered: 25, 50 and 100 generations.
A smaller value of $\tau_t$ indicates a higher frequency of environmental change.
\item Each algorithm is executed independently 31 times to ensure statistical reliability.
\item Before the first environmental change, each algorithm is allowed 300 generations to enable sufficient convergence of the initial population.
\end{itemize}

\subsection{Performance Metric}

The following widely used performance metric for dynamic multi-objective optimization is employed~\cite{li2007metrics,jiang2023survey}.

\textbf{Mean Hypervolume (MHV)} measures the average hypervolume of the obtained solution set over all time steps, reflecting both convergence and diversity.
The hypervolume indicator is widely used in multi-objective optimization because it can simultaneously capture convergence and spread of approximation sets~\cite{zitzler1999multiobjective,while2006faster}. 

Let ${S}_t$ denote the set of objective vectors obtained
by an algorithm at time step $t$, $PF_t$ denotes
the true Pareto front at time step $t$, and $T$ is the set of discrete time steps during one run. Formally, MHV is defined as
\begin{equation}
\mathrm{MHV} =
\frac{1}{|T|}
\sum_{t \in T}
{\mathrm{HV}(\widehat{S}_t)},
\end{equation}
where $\widehat{S}_t$ is the normalized set of $S_t$ based on $PF_t$, i.e., 
for each $\mathbf{f}\in S_t$, the corresponding $\widehat{\mathbf{f}}\in \widehat{S}_t$ is obtained by
\begin{equation}
\widehat{f}_i =
\frac{f_i - z^*_i(t)}
{z^{\mathrm{nad}}_i(t) - z^*_i(t)},
\quad i = 1, \ldots, m(t),
\end{equation}
where $\mathbf{z}^*(t) = (z^*_1(t), \ldots, z^*_{m(t)}(t))^\top$ and
$\mathbf{z}^{\mathrm{nad}}(t) = (z^{\mathrm{nad}}_1(t), \ldots, z^{\mathrm{nad}}_{m(t)}(t))^\top$
denote the ideal point and nadir point of the true Pareto front $PF_t$,
respectively.

However, the hypervolume
value $\mathrm{HV}(\widehat{S}_t)$ is strongly influenced by the
dimensionality $m(t)$ of the objective space.
Consequently, $\mathrm{HV}(\widehat{S}_t)$ may have substantially different
numerical scales at different time steps. Therefore, if the mean hypervolume were computed directly as
$\frac{1}{|T|}\sum_{t \in T} \mathrm{HV}(\widehat{S}_t)$, time steps associated
with larger hypervolume values would dominate the average, while those with
smaller hypervolume values would be under-represented.
This would result in a biased performance assessment in dynamic scenarios with a
changing number of objectives.

In this paper, we slightly modify the definition of MHV as follows:
\begin{equation}
\mathrm{MHV} =
\frac{1}{|T|}
\sum_{t \in T}
\frac{\mathrm{HV}(\widehat{S}_t)}
{\mathrm{HV}(\widehat{PF}_t)},
\end{equation}
where $\widehat{PF}_t$ is the normalized Pareto front at time step $t$.

The hypervolume in the normalized objective space is computed as
\begin{equation}
\mathrm{HV}(\widehat{A}) =
\mathrm{VOL}
\left(
\bigcup_{\widehat{\mathbf{f}} \in \widehat{A}}
\left[
\widehat{f}_1, 1.1
\right]
\times \cdots \times
\left[
\widehat{f}_{m(t)}, 1.1
\right]
\right),
\end{equation}
where $\widehat{A} \in \{\widehat{S}_t, \widehat{PF}_t\}$ and
$\mathrm{VOL}(\cdot)$ denotes the Lebesgue measure.
Objective vectors outside the hyper-rectangle $[0,1.1]^{m(t)}$ are discarded
before the hypervolume calculation.

By normalizing $\mathrm{HV}(\widehat{S}_t)$ with respect to
$\mathrm{HV}(\widehat{PF}_t)$, which represents the maximum achievable
hypervolume at time step $t$, the resulting values are mapped to a more
comparable scale across different dimensions.
This modification is intended to reduce the bias introduced by different
objective dimensions when aggregating dynamic performance over time.


\subsection{Results and Discussion}

Tables~\ref{tab:mhv_results_set1}--\ref{tab:mhv_results_set3} report the MHV values under the three dynamic settings, while Figs.~\ref{fig:mhv_setting1}--\ref{fig:mhv_setting3} show the corresponding Friedman rankings over all test problems.
Overall, the proposed benchmark produces a clear separation among the compared algorithms.
KTDMOEA achieves the best overall performance on most test instances, especially when the change frequency is moderate or low ($\tau_t=50$ or $100$), while STA is the second strongest method and becomes particularly competitive under severe changes.
By contrast, DTAEA is mainly competitive in a limited set of rapidly changing cases, and LEC is consistently inferior on most instances.
These results suggest that the proposed benchmark can reveal not only absolute performance differences, but also different adaptation characteristics of algorithmic mechanisms.

\subsubsection{Performance Under Mild Objective Changes}

\begin{figure*}[!htb]
\centering
\subfigure[$\tau_t=25$]{
    \includegraphics[width=0.3\textwidth]{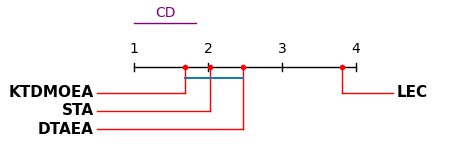}
    \label{fig:setting1_tau25}
}
\hfill
\subfigure[$\tau_t=50$]{
    \includegraphics[width=0.3\textwidth]{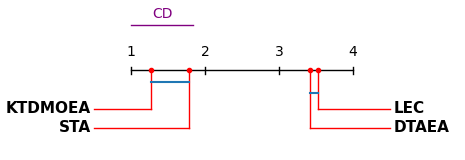}
    \label{fig:setting1_tau50}
}
\hfill
\subfigure[$\tau_t=100$]{
    \includegraphics[width=0.3\textwidth]{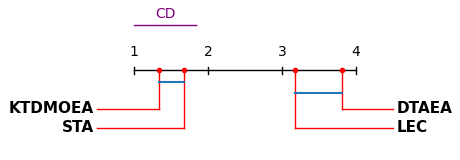}
    \label{fig:setting1_tau100}
}
\caption{Friedman ranking among MHV of obtained solutions  by four algorithms under \textbf{Setting I}.}
\label{fig:mhv_setting1}
\end{figure*}

The results under \textbf{Setting I} show that even mild objective additions/removals are sufficient to generate meaningful differences among algorithms.
When $\tau_t=25$, DTAEA still wins several Minus-DTLZ instances such as Minus-DTLZ2--Minus-DTLZ4, indicating that its dual-archive mechanism is effective when the environment changes so frequently that more sophisticated transfer or learning mechanisms cannot be fully exploited.
However, as $\tau_t$ increases, the advantage shifts clearly to KTDMOEA and, in a few cases, STA.
For example, on Minus-DTLZ2, Minus-DTLZ3, Minus-WFG4--Minus-WFG6, and Minus-WFG8, KTDMOEA improves substantially from $\tau_t=25$ to $\tau_t=100$ and becomes the best method by a clear margin.
This trend is also reflected by the Friedman rankings in Fig.~\ref{fig:mhv_setting1}, where KTDMOEA and STA occupy the top two positions for $\tau_t=50$ and $100$, while DTAEA is comparatively more competitive only in the fastest-changing case.

This trend indicates that mild changes preserve substantial structural continuity between consecutive environments.
As a result, transferred knowledge remains useful after objective addition or removal, provided that the algorithm is given enough generations to refine the transferred solutions.

\begin{table*}[!htb]
\centering
\caption{MHV results (mean $\pm$ standard deviation) obtained by four algorithms under \textbf{Setting I}.}
\label{tab:mhv_results_set1}
\fontsize{8pt}{10pt}\selectfont
\begin{tabular}{c c c c c c}
\hline
{Problem} & {$\tau_t$} 
& DTAEA & KTDMOEA & LEC & STA \\
\hline

\multirow{3}{*}{Minus-DTLZ1}& 25   & $0.5757 (4.45e-03)$& $\textbf{0.6080} (1.80e-02)$& $0.4769 (2.79e-02)$& $0.5657 (1.91e-02)$\\
& 50  & $0.5727 (3.93e-03)$& $\textbf{0.6728} (1.22e-02)$& $0.5784 (1.85e-02)$& $0.6561 (1.57e-02)$\\
& 100  & $0.5621 (3.10e-03)$& $\textbf{0.7095} (8.09e-03)$& $0.6359 (1.07e-02)$& $0.7007 (7.23e-03)$\\
\hline

\multirow{3}{*}{Minus-DTLZ2}& 25   & $\textbf{0.7226} (1.01e-02)$& $0.7060 (2.25e-02)$& $0.5159 (3.95e-02)$& $0.7167 (2.34e-02)$\\
& 50  & $0.6784 (5.02e-03)$& $\textbf{0.8195} (1.09e-02)$& $0.6556 (2.39e-02)$& $0.8043 (1.22e-02)$\\
& 100  & $0.6581 (3.14e-03)$& $\textbf{0.8598} (4.51e-03)$& $0.7595 (2.48e-02)$& $0.8568 (5.00e-03)$\\
\hline

\multirow{3}{*}{Minus-DTLZ3}& 25   & $\textbf{0.6927} (6.54e-03)$& $0.6769 (2.75e-02)$& $0.5620 (3.86e-02)$& $0.6471 (2.54e-02)$\\
& 50  & $0.6829 (4.28e-03)$& $\textbf{0.7935} (1.53e-02)$& $0.7027 (2.14e-02)$& $0.7626 (1.22e-02)$\\
& 100  & $0.6627 (2.43e-03)$& $\textbf{0.8352} (4.96e-03)$& $0.7841 (1.01e-02)$& $0.8266 (6.69e-03)$\\
\hline

 \multirow{3}{*}{Minus-DTLZ4}& 25  & $\textbf{0.6775} (1.06e-02)$& $0.6083 (7.83e-02)$& $0.4013 (6.84e-02)$& $0.5453 (7.89e-02)$\\
 & 50  & $0.6842 (5.94e-03)$& $0.7422 (1.01e-01)$& $0.5058 (8.75e-02)$& $\textbf{0.7491} (8.46e-02)$\\
 & 100 & $0.6574 (4.04e-03)$& $\textbf{0.8377} (2.97e-02)$& $0.6573 (5.84e-02)$& $0.8226 (6.01e-02)$\\
\hline

\multirow{3}{*}{Minus-WFG4}& 25   & $0.7271 (5.98e-03)$& $\textbf{0.8182} (9.57e-03)$& $0.7144 (2.95e-02)$& $0.7920 (1.44e-02)$\\
& 50  & $0.7077 (4.10e-03)$& $\textbf{0.8208} (6.49e-03)$& $0.7467 (2.62e-02)$& $0.8132 (5.02e-03)$\\
& 100  & $0.6809 (4.92e-03)$& $\textbf{0.8140} (5.11e-03)$& $0.7626 (1.31e-02)$& $0.8114 (5.48e-03)$\\
\hline

\multirow{3}{*}{Minus-WFG5}& 25   & $0.7125 (4.96e-03)$& $\textbf{0.8163} (1.11e-02)$& $0.7471 (1.87e-02)$& $0.7844 (1.25e-02)$\\
& 50  & $0.6924 (3.78e-03)$& $\textbf{0.8252} (3.64e-03)$& $0.7718 (1.49e-02)$& $0.8088 (8.05e-03)$\\
& 100  & $0.6764 (3.70e-03)$& $\textbf{0.8251} (4.78e-03)$& $0.7868 (1.18e-02)$& $0.8195 (4.77e-03)$\\
\hline

\multirow{3}{*}{Minus-WFG6}& 25   & $0.7088 (5.97e-03)$& $\textbf{0.7201} (3.15e-02)$& $0.5157 (3.12e-02)$& $0.7104 (2.30e-02)$\\
& 50  & $0.6881 (5.07e-03)$& $\textbf{0.8183} (1.27e-02)$& $0.6343 (2.61e-02)$& $0.7913 (1.39e-02)$\\
& 100  & $0.6704 (4.73e-03)$& $\textbf{0.8406} (5.25e-03)$& $0.7504 (2.56e-02)$& $0.8362 (5.65e-03)$\\
\hline

 \multirow{3}{*}{Minus-WFG7}& 25  & $0.6984 (6.24e-03)$& $\textbf{0.7405} (3.78e-02)$& $0.4391 (4.20e-02)$& $0.7334 (3.42e-02)$\\
 & 50  & $0.6777 (4.31e-03)$& $\textbf{0.7991} (1.48e-02)$& $0.5833 (4.03e-02)$& $0.7962 (1.88e-02)$\\
 & 100 & $0.6634 (4.84e-03)$& $0.8060 (1.19e-02)$& $0.6678 (3.03e-02)$& $\textbf{0.8143} (5.36e-03)$\\
\hline

\multirow{3}{*}{Minus-WFG8}& 25   & $0.7114 (5.96e-03)$& $0.7424 (2.60e-02)$& $0.4366 (4.02e-02)$& $\textbf{0.7784} (1.86e-02)$\\
& 50  & $0.6895 (4.76e-03)$& $\textbf{0.8445} (1.03e-02)$& $0.5320 (3.98e-02)$& $0.8177 (1.24e-02)$\\
& 100  & $0.6817 (4.05e-03)$& $\textbf{0.8775} (7.26e-03)$& $0.6212 (4.24e-02)$& $0.8599 (4.93e-03)$\\
\hline

 \multirow{3}{*}{Minus-WFG9}& 25  & $0.7499 (6.01e-03)$& $0.8488 (3.38e-02)$& $0.7445 (2.25e-02)$& $\textbf{0.8639} (8.42e-03)$\\
 & 50  & $0.7223 (5.41e-03)$& $0.8565 (1.98e-02)$& $0.7977 (2.01e-02)$& $\textbf{0.8696} (6.77e-03)$\\
 & 100 & $0.6996 (4.82e-03)$& $0.8624 (1.09e-02)$& $0.8070 (1.45e-02)$& $\textbf{0.8671} (4.78e-03)$\\
\hline

\end{tabular}
\end{table*}

\subsubsection{Performance Under Moderate Objective Changes}

\begin{figure*}[!htb]
\centering
\subfigure[$\tau_t=25$]{
    \includegraphics[width=0.3\textwidth]{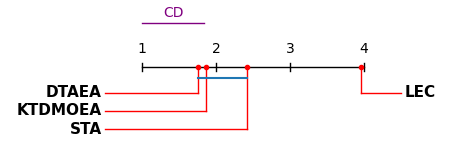}
    \label{fig:setting2_tau25}
}
\hfill
\subfigure[$\tau_t=50$]{
    \includegraphics[width=0.3\textwidth]{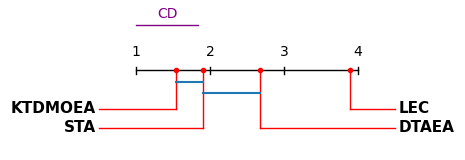}
    \label{fig:setting2_tau50}
}
\hfill
\subfigure[$\tau_t=100$]{
    \includegraphics[width=0.3\textwidth]{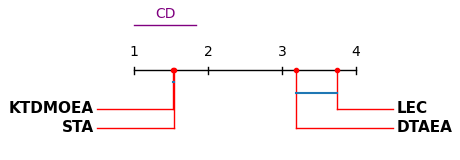}
    \label{fig:setting2_tau100}
}
\caption{Friedman ranking among MHV of obtained solutions  by four algorithms under \textbf{Setting II}.}
\label{fig:mhv_setting2}
\end{figure*}

Under \textbf{Setting II}, where two objectives are added or removed at each change, the relative ranking becomes more stable and the superiority of KTDMOEA and STA becomes more evident.
Compared with Setting I, the MHV values of all algorithms decrease on many problems, especially on Minus-DTLZ1, Minus-DTLZ4, Minus-WFG7, and Minus-WFG8, showing that larger jumps in the number of objectives substantially increase the difficulty of post-change recovery.
Nevertheless, KTDMOEA still dominates most instances, whereas STA obtains the best performance on several difficult cases such as Minus-DTLZ1 ($\tau_t=100$), Minus-DTLZ4 ($\tau_t=100$), Minus-WFG7 ($\tau_t=50,100$), and Minus-WFG9 ($\tau_t=50,100$).
The same pattern can be observed in Fig.~\ref{fig:mhv_setting2}, where KTDMOEA attains the best overall rank in most cases and STA remains close behind, confirming that these two methods are more robust than DTAEA and LEC when the dimensional jump becomes larger.

These results show that the proposed benchmark does not merely rank algorithms by overall search ability.
Instead, it exposes different adaptation biases: KTDMOEA is stronger when cross-environment knowledge remains reusable after the change, whereas STA becomes more attractive when the dimensional jump weakens the reliability of direct transfer.

\begin{table*}[!htb]
\centering
\caption{MHV results (mean $\pm$ standard deviation) obtained by four algorithms under \textbf{Setting II}.}
\label{tab:mhv_results_set2} 
\fontsize{8pt}{10pt}\selectfont
\begin{tabular}{c c c c c c}
\hline
{Problem} & {$\tau_t$} 
& DTAEA & KTDMOEA & LEC & STA \\
\hline

\multirow{3}{*}{Minus-DTLZ1}& 25   & $0.3573 (9.32e-03)$& $\textbf{0.4403} (1.80e-02)$& $0.2973 (2.16e-02)$& $0.3935 (3.22e-02)$\\
& 50  & $0.3749 (8.47e-03)$& $\textbf{0.4827} (3.03e-02)$& $0.3703 (1.23e-02)$& $0.4748 (2.74e-02)$\\
& 100  & $0.3776 (1.26e-02)$& $0.4892 (3.09e-02)$& $0.4088 (1.19e-02)$& $\textbf{0.5039} (2.91e-02)$\\
\hline

\multirow{3}{*}{Minus-DTLZ2}& 25   & $\textbf{0.6818} (2.35e-02)$& $0.6561 (2.16e-02)$& $0.3469 (2.04e-02)$& $0.6376 (2.75e-02)$\\
& 50  & $0.6457 (1.40e-02)$& $\textbf{0.7898} (2.44e-02)$& $0.4657 (2.67e-02)$& $0.7699 (2.40e-02)$\\
& 100  & $0.5954 (1.30e-02)$& $\textbf{0.8211} (2.07e-02)$& $0.5808 (2.38e-02)$& $0.8127 (2.59e-02)$\\
\hline

\multirow{3}{*}{Minus-DTLZ3}& 25   & $0.6068 (2.42e-02)$& $\textbf{0.6072} (2.22e-02)$& $0.3760 (2.02e-02)$& $0.5289 (3.12e-02)$\\
& 50  & $0.6529 (1.93e-02)$& $\textbf{0.7422} (2.29e-02)$& $0.5153 (2.12e-02)$& $0.6874 (2.02e-02)$\\
& 100  & $0.6098 (1.64e-02)$& $\textbf{0.7659} (2.06e-02)$& $0.6119 (2.40e-02)$& $0.7532 (2.10e-02)$\\
\hline

 \multirow{3}{*}{Minus-DTLZ4}& 25  & $\textbf{0.5448} (2.84e-02)$& $0.4085 (6.21e-02)$& $0.2645 (4.40e-02)$& $0.3430 (5.09e-02)$\\
 & 50  & $\textbf{0.6466} (1.91e-02)$& $0.5497 (7.70e-02)$& $0.3166 (2.58e-02)$& $0.6222 (7.16e-02)$\\
 & 100 & $0.5594 (1.22e-02)$& $0.7161 (4.84e-02)$& $0.3749 (3.69e-02)$& $\textbf{0.7417} (5.42e-02)$\\
\hline

\multirow{3}{*}{Minus-WFG4}& 25   & $0.6972 (2.07e-02)$& $\textbf{0.7026} (3.14e-02)$& $0.5316 (3.32e-02)$& $0.6762 (2.31e-02)$\\
& 50  & $\textbf{0.7008} (1.67e-02)$& $0.6930 (2.04e-02)$& $0.5335 (3.68e-02)$& $0.6772 (1.63e-02)$\\
& 100  & $0.6496 (1.92e-02)$& $\textbf{0.6724} (1.96e-02)$& $0.4899 (2.49e-02)$& $0.6640 (2.11e-02)$\\
\hline

\multirow{3}{*}{Minus-WFG5}& 25   & $0.6716 (2.09e-02)$& $\textbf{0.7072} (2.59e-02)$& $0.5572 (2.63e-02)$& $0.6663 (2.48e-02)$\\
& 50  & $0.6531 (1.25e-02)$& $\textbf{0.7183} (2.25e-02)$& $0.5832 (2.54e-02)$& $0.6899 (2.66e-02)$\\
& 100  & $0.6139 (1.46e-02)$& $\textbf{0.7046} (1.54e-02)$& $0.5741 (2.76e-02)$& $0.6911 (2.19e-02)$\\
\hline

\multirow{3}{*}{Minus-WFG6}& 25   & $\textbf{0.6301} (2.42e-02)$& $0.6075 (2.28e-02)$& $0.3490 (2.16e-02)$& $0.5532 (4.09e-02)$\\
& 50  & $0.6367 (1.89e-02)$& $\textbf{0.6947} (2.37e-02)$& $0.4554 (2.12e-02)$& $0.6628 (1.96e-02)$\\
& 100  & $0.5854 (1.33e-02)$& $\textbf{0.7025} (1.80e-02)$& $0.5654 (2.90e-02)$& $0.6860 (1.90e-02)$\\
\hline

 \multirow{3}{*}{Minus-WFG7}& 25  & $\textbf{0.7029} (2.67e-02)$& $0.5552 (4.50e-02)$& $0.2832 (1.70e-02)$& $0.5774 (2.74e-02)$\\
 & 50  & $0.6499 (1.72e-02)$& $0.6366 (5.11e-02)$& $0.3505 (1.72e-02)$& $\textbf{0.6692} (2.38e-02)$\\
 & 100 & $0.6049 (1.25e-02)$& $0.6601 (2.43e-02)$& $0.3975 (2.33e-02)$& $\textbf{0.6777} (1.91e-02)$\\
\hline

\multirow{3}{*}{Minus-WFG8}& 25   & $\textbf{0.7544} (2.16e-02)$& $0.7076 (2.51e-02)$& $0.3190 (3.06e-02)$& $0.7232 (3.77e-02)$\\
& 50  & $0.6484 (2.03e-02)$& $\textbf{0.8434} (3.18e-02)$& $0.3990 (3.96e-02)$& $0.7965 (2.23e-02)$\\
& 100  & $0.5855 (1.22e-02)$& $\textbf{0.8400} (3.21e-02)$& $0.4866 (6.02e-02)$& $0.8194 (2.76e-02)$\\
\hline

 \multirow{3}{*}{Minus-WFG9}& 25  & $0.6115 (2.30e-02)$& $0.7182 (3.33e-02)$& $0.5786 (5.33e-02)$& $\textbf{0.7184} (2.99e-02)$\\
 & 50  & $0.6129 (2.48e-02)$& $0.7355 (5.55e-02)$& $0.6210 (4.77e-02)$& $\textbf{0.7371} (2.64e-02)$\\
 & 100 & $0.6144 (3.09e-02)$& $0.7177 (2.97e-02)$& $0.5836 (3.80e-02)$& $\textbf{0.7310} (2.26e-02)$\\
\hline
\end{tabular}
\end{table*}

\subsubsection{Performance Under Severe and Irregular Changes}

\begin{figure*}[!htb]
\centering
\subfigure[$\tau_t=25$]{
    \includegraphics[width=0.3\textwidth]{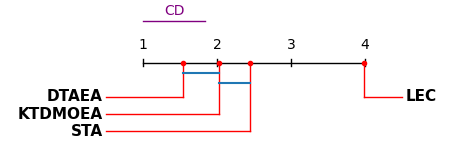}
    \label{fig:setting3_tau25}
}
\hfill
\subfigure[$\tau_t=50$]{
    \includegraphics[width=0.3\textwidth]{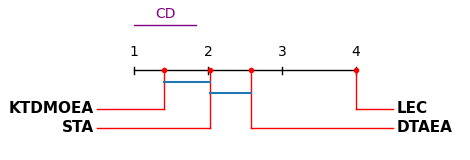}
    \label{fig:setting3_tau50}
}
\hfill
\subfigure[$\tau_t=100$]{
    \includegraphics[width=0.3\textwidth]{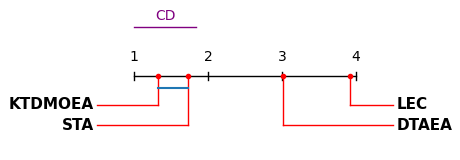}
    \label{fig:setting3_tau100}
}
\caption{Friedman ranking among MHV of obtained solutions  by four algorithms under \textbf{Setting III}.}
\label{fig:mhv_setting3}
\end{figure*}

The distinction becomes even clearer in \textbf{Setting III}, which involves irregular and severe changes in the number of objectives.
Here, STA wins all three Minus-DTLZ1 cases and the two larger-$\tau_t$ cases of Minus-DTLZ4, demonstrating strong robustness on landscapes where both the dimensionality and the active objective composition vary abruptly.
At the same time, KTDMOEA still achieves the best results on most Minus-DTLZ2, Minus-DTLZ3, and Minus-WFG instances once $\tau_t \geq 50$.
This indicates that severe changes do not uniformly invalidate knowledge transfer; rather, the usefulness of historical information depends on whether the geometry of the Pareto set remains sufficiently correlated after the objective subset changes.
Figure~\ref{fig:mhv_setting3} shows a clear separation between KTDMOEA/STA and the other two algorithms, especially when $\tau_t=50$ and $100$.

DTAEA is occasionally the best method at $\tau_t=25$ on problems such as Minus-DTLZ2, Minus-DTLZ3, Minus-DTLZ4, Minus-WFG6, Minus-WFG7, and Minus-WFG8, but this advantage diminishes quickly as more search time becomes available.
This suggests that DTAEA is effective at short-term response immediately after an environmental change, but it is less competitive than KTDMOEA and STA in sustained re-adaptation.

\begin{table*}[!htb]
\centering
\caption{MHV results (mean $\pm$ standard deviation) obtained by four algorithms under \textbf{Setting III}.}
\label{tab:mhv_results_set3} 
\fontsize{8pt}{10pt}\selectfont
\begin{tabular}{c c c c c c}
\hline
{Problem} & {$\tau_t$} 
& DTAEA & KTDMOEA & LEC & STA \\
\hline

\multirow{3}{*}{Minus-DTLZ1}& 25   & $0.2895 (1.26e-02)$& $0.3115 (3.09e-02)$& $0.2257 (1.41e-02)$& $\textbf{0.3340} (4.60e-02)$\\
& 50  & $0.3346 (1.94e-02)$& $0.3800 (1.96e-02)$& $0.2679 (9.69e-03)$& $\textbf{0.4066} (6.54e-02)$\\
& 100  & $0.3402 (1.71e-02)$& $0.4192 (2.06e-02)$& $0.3211 (1.03e-02)$& $\textbf{0.4357} (5.29e-02)$\\
\hline

\multirow{3}{*}{Minus-DTLZ2}& 25   & $\textbf{0.6194} (1.77e-02)$& $0.5351 (3.27e-02)$& $0.2826 (2.33e-02)$& $0.5648 (2.78e-02)$\\
& 50  & $0.6056 (1.65e-02)$& $\textbf{0.7254} (2.09e-02)$& $0.3520 (2.18e-02)$& $0.7101 (2.62e-02)$\\
& 100  & $0.5458 (1.40e-02)$& $\textbf{0.7798} (2.69e-02)$& $0.4546 (1.99e-02)$& $0.7621 (1.82e-02)$\\
\hline

\multirow{3}{*}{Minus-DTLZ3}& 25   & $\textbf{0.5216} (3.20e-02)$& $0.4577 (2.22e-02)$& $0.3041 (1.81e-02)$& $0.4494 (3.10e-02)$\\
& 50  & $0.5703 (1.58e-02)$& $\textbf{0.6581} (2.51e-02)$& $0.4039 (1.73e-02)$& $0.6232 (2.84e-02)$\\
& 100  & $0.5546 (1.11e-02)$& $\textbf{0.7222} (2.12e-02)$& $0.5206 (1.37e-02)$& $0.6953 (2.90e-02)$\\
\hline

 \multirow{3}{*}{Minus-DTLZ4}& 25  & $\textbf{0.4995} (1.82e-02)$& $0.3903 (4.72e-02)$& $0.1991 (4.81e-02)$& $0.2871 (3.99e-02)$\\
 & 50  & $0.5926 (2.30e-02)$& $0.5624 (6.73e-02)$& $0.2622 (3.89e-02)$& $\textbf{0.5966} (5.64e-02)$\\
 & 100 & $0.5394 (1.92e-02)$& $0.6824 (5.30e-02)$& $0.3399 (1.74e-02)$& $\textbf{0.7231} (3.81e-02)$\\
\hline

\multirow{3}{*}{Minus-WFG4}& 25   & $0.6043 (3.61e-02)$& $\textbf{0.6386} (2.47e-02)$& $0.3784 (2.40e-02)$& $0.5664 (3.33e-02)$\\
& 50  & $0.6388 (1.83e-02)$& $\textbf{0.6612} (2.68e-02)$& $0.4197 (2.50e-02)$& $0.6247 (2.35e-02)$\\
& 100  & $0.5981 (1.98e-02)$& $\textbf{0.6396} (2.62e-02)$& $0.4445 (3.05e-02)$& $0.5992 (1.88e-02)$\\
\hline

\multirow{3}{*}{Minus-WFG5}& 25   & $0.5959 (1.84e-02)$& $\textbf{0.6192} (2.86e-02)$& $0.4462 (2.68e-02)$& $0.5785 (2.55e-02)$\\
& 50  & $0.6066 (1.83e-02)$& $\textbf{0.6711} (2.34e-02)$& $0.4765 (2.63e-02)$& $0.6115 (2.23e-02)$\\
& 100  & $0.5608 (1.41e-02)$& $\textbf{0.6555} (2.23e-02)$& $0.5116 (2.05e-02)$& $0.6292 (2.56e-02)$\\
\hline

\multirow{3}{*}{Minus-WFG6}& 25   & $\textbf{0.5357} (1.78e-02)$& $0.5128 (2.08e-02)$& $0.2750 (1.83e-02)$& $0.4653 (2.26e-02)$\\
& 50  & $0.5941 (2.15e-02)$& $\textbf{0.6580} (2.14e-02)$& $0.3441 (1.96e-02)$& $0.5808 (2.37e-02)$\\
& 100  & $0.5368 (1.45e-02)$& $\textbf{0.6681} (2.33e-02)$& $0.4249 (2.45e-02)$& $0.6199 (1.92e-02)$\\
\hline

 \multirow{3}{*}{Minus-WFG7}& 25  & $\textbf{0.6361} (2.27e-02)$& $0.5171 (4.26e-02)$& $0.2212 (2.47e-02)$& $0.5093 (3.37e-02)$\\
 & 50  & $0.6109 (2.18e-02)$& $\textbf{0.6284} (3.45e-02)$& $0.2752 (2.07e-02)$& $0.6101 (2.29e-02)$\\
 & 100 & $0.5463 (1.54e-02)$& $\textbf{0.6480} (2.20e-02)$& $0.3386 (1.90e-02)$& $0.6206 (2.40e-02)$\\
\hline

\multirow{3}{*}{Minus-WFG8}& 25   & $\textbf{0.7775} (2.90e-02)$& $0.5730 (3.22e-02)$& $0.2543 (2.74e-02)$& $0.6114 (3.83e-02)$\\
& 50  & $0.6317 (1.93e-02)$& $\textbf{0.7906} (2.20e-02)$& $0.2996 (3.17e-02)$& $0.7266 (2.94e-02)$\\
& 100  & $0.5410 (1.94e-02)$& $\textbf{0.7826} (3.33e-02)$& $0.3542 (4.21e-02)$& $0.7674 (3.32e-02)$\\
\hline

 \multirow{3}{*}{Minus-WFG9}& 25  & $0.5883 (2.85e-02)$& $\textbf{0.6168} (4.99e-02)$& $0.4479 (3.57e-02)$& $0.6011 (3.31e-02)$\\
 & 50  & $0.5682 (1.93e-02)$& $\textbf{0.6806} (3.96e-02)$& $0.4893 (3.60e-02)$& $0.6603 (3.03e-02)$\\
 & 100 & $0.5134 (2.07e-02)$& $\textbf{0.6885} (2.90e-02)$& $0.5224 (2.82e-02)$& $0.6708 (2.63e-02)$\\
\hline
\end{tabular}
\end{table*}

\subsubsection{Effect of Change Frequency}

Another notable observation is the strong interaction between change frequency and algorithm mechanism.
For KTDMOEA, performance generally improves as $\tau_t$ increases, often substantially.
This pattern is especially evident on Minus-DTLZ2, Minus-DTLZ3, Minus-WFG6, Minus-WFG8, and Minus-WFG9 across all three settings.
The reason is intuitive: knowledge transfer mainly provides a better initial state after each change, but its full benefit can only be realized if the population is allowed enough generations to refine the transferred solutions under the new objective subset.

STA exhibits a similar but slightly milder trend, indicating that its specialized adaptation mechanism also benefits from additional recovery time.
In contrast, DTAEA frequently degrades as $\tau_t$ increases, which implies that its main strength lies in maintaining a survival-ready population for rapid environmental response rather than in progressively exploiting environment-specific structure.
Hence, the frequency parameter $\tau_t$ does not only control problem difficulty; it also determines which adaptation capability is emphasized by the benchmark.

\subsubsection{Implications for Benchmark Design}

From a benchmark-design perspective, the above results confirm the usefulness of the proposed test suite.
The benchmark generates not only different levels of difficulty, but also qualitatively different adaptation scenarios: gradual changes favor reusable knowledge, severe irregular changes reward robust restructuring ability, and high-frequency changes emphasize immediate response capacity.
More importantly, since the objective functions remain fixed and only the active objective subset changes, the observed performance differences can be interpreted directly in terms of how well an algorithm handles objective addition and removal.
This makes the proposed benchmark more behaviorally informative than formulations in which the objective definitions change implicitly together with the number of objectives.




\section{Conclusion}

This paper analyzed a fundamental limitation of existing benchmark test suites for dynamic multi-objective optimization with a changing number of objectives.
We showed that commonly used benchmarks unintentionally change objective functions over time, which contradicts the intended problem definition.

To address this issue, we proposed a scalable benchmark test suite in which objective functions remain fixed and only the number of active objectives changes over time.
Our framework is flexible, theoretically consistent, and applicable to a wide range of dynamic scenarios.
Benchmark studies using representative algorithms demonstrate the usefulness of the proposed test suite.
For reproducibility, the benchmark suite and the implementations of the compared algorithms used in this study are publicly available at \url{https://github.com/MOL-SZU/DMOP}.

In future work, we plan to develop new algorithms specifically designed to exploit the structural properties of dynamic objective addition and removal.

\bibliographystyle{splncs04}
\bibliography{references}
\end{document}